%% file: main.tex
\documentclass{article}

\usepackage{iclr2027_conference,times}
\input{math_commands.tex}

\usepackage{amsmath}
\usepackage{amssymb}
\usepackage{booktabs}
\usepackage{float}
\usepackage{graphicx}
\usepackage{makecell}
\usepackage{microtype}
\usepackage{multirow}
\usepackage{placeins}
\usepackage{wrapfig}
\usepackage{xcolor}
\usepackage{colortbl}
\usepackage{xspace}
\usepackage{tcolorbox}
\tcbuselibrary{listings,skins}
\usepackage{url}
\usepackage{hyperref}

\definecolor{herocyan}{RGB}{66,196,188}
\definecolor{heropurple}{RGB}{116,76,224}
\definecolor{herogray}{RGB}{242,242,244}

\definecolor{promptbg}{gray}{0.965}
\definecolor{promptframe}{gray}{0.72}
\definecolor{promptlabel}{RGB}{70,100,150}
\newtcblisting{promptbox}{
  enhanced,
  listing only,
  listing options={basicstyle=\ttfamily\scriptsize,breaklines=true,
    breakindent=0pt,columns=fullflexible,keepspaces=true},
  colback=promptbg,
  colframe=promptframe,
  boxrule=0.3pt,
  arc=1pt,
  left=4pt,right=4pt,top=2pt,bottom=2pt
}
\newcommand{\promptrole}[1]{{\footnotesize\sffamily\bfseries\color{promptlabel}#1}\par\vspace{1pt}}

\title{HeroFrame-Bench: Reference-Anchored Evaluation via Rubric--Ranking
Co-Evolution for Movie Hero Frame Selection}

\author{
Weitai Kang\textsuperscript{1},
Hanieh Deilamsalehy\textsuperscript{2},
Yumo Xu\textsuperscript{2},
Dewang Sultania\textsuperscript{2},
Serdar Cellat\textsuperscript{2},
Yan Yan\textsuperscript{1}
\\
\textsuperscript{1}University of Illinois Chicago
\quad
\textsuperscript{2}Netflix
}

\iclrfinalcopy

\begin{document}

\maketitle

\begin{abstract}
\input{sections/00_abstract}
\end{abstract}

\input{sections/01_introduction}

\input{sections/02_related_work}
\input{sections/03_method}
\input{sections/04_experiments}
\input{sections/05_conclusion}

\input{sections/06_ai_use_statement}
\input{sections/07_ethics_statement}
\input{sections/08_reproducibility_statement}

\bibliography{references}
\bibliographystyle{iclr2027_conference}

\appendix
\input{sections/09_appendix}

\end{document}

%% file: math_commands.tex
\usepackage{amsmath,amsfonts,bm}

\def\eqref#1{equation~\ref{#1}}
\def\1{\bm{1}}

\DeclareMathAlphabet{\mathsfit}{\encodingdefault}{\sfdefault}{m}{sl}
\SetMathAlphabet{\mathsfit}{bold}{\encodingdefault}{\sfdefault}{bx}{n}

%% file: sections/00_abstract.tex
Hero frames are in-film stills used as source imagery for theatrical posters,
streaming cover art, film database listings, and other promotional placements.
As the first visual entry point, they shape audiences' initial
impressions of the movie and their subsequent willingness to watch it.
Selecting these frames, a task we term \emph{hero frame selection}, requires
balancing content relevance with aesthetic appeal.
A related task is keyframe selection, yet its benchmarks prioritize relevance
over aesthetics, using either finite
annotations that exclude valid alternatives or VideoQA that
entangles selection quality with downstream model capability.
We therefore introduce
\textbf{\textsc{HeroFrame-Bench}}, built through a scalable \emph{VLM-as-a-Judge} framework. 
We construct multimodal contexts from diverse metadata to ground
a VLM judge that scores selected frames using our
\emph{Reference-anchored Percentile}. The percentile is obtained by inserting
each frame into reusable, pre-ranked reference chains, enabling direct,
extensible, and reliable evaluation. To reduce ambiguity and improve consistency
in these subjective judgements, we further propose
\emph{Rubric--Ranking Co-Evolution},
which generates movie-specific rubrics to condition the VLM judge and refines
rubrics jointly with the resulting rankings. Within
this process, we introduce several verifiable signals, most notably the
\emph{Inverted Rubric Attack}, to select robust rubrics.
Finally, \textsc{HeroFrame-Bench} is instantiated over 204 movies with 2,031 reference
chains and 1,970 learned rubrics. We build an annotation interface for
human-alignment studies which show that our construction design improves VLM agreement with human from 77.56\% to
83.78\%. Evaluation on multiple methods show that hero frame selection
remains challenging.

%% file: sections/01_introduction.tex
\section{Introduction}

Prominent imagery is integral to how film databases~\citep{imdb2026primary},
theatrical promotion~\citep{gracewood2022still}, and streaming services
~\citep{netflix2018ava,amazon2026artwork,apple2026hero,roku2026hero,
eklund2022thumbnails,ceuterick2024thumbnails,vanes2025interface} present and
promote movies and other media content by communicating subject matter,
capturing audience attention, and influencing viewing decisions
~\citep{netflix2016picture}. A \emph{hero frame} is an in-film still (frame) selected
for this role, requiring not only movie-specific content relevance but also
aesthetic appeal~\citep{pond52021hero,song2016click}. 
Automatically selecting such a hero frame given a video, which we term \emph{hero frame selection}, therefore constitutes an important video understanding task that remains understudied. 

% \yumo{Hero frame is an entity, while hero frame selection is a task. Currently, 1st paragraph introduces hero frame (an entity), while 2nd paragraph talks about KFS (a task), so the transition feels slightly jumpy. Also, if hero frame selection is a new task and we are the first to propose studying it, we should highlight that: we believe it's an important task, but it's understudied due to the challenges in evaluation (now at the end of the 2nd paragraph; consider bring it forward here as you see fit): \textit{joint assessment of movie-specific content relevance, aesthetic appeal, and their interaction.} $\longrightarrow$ then 2nd paragraph contrasts it with KFS $\longrightarrow$ 3rd paragraph for our methodology}

% A related task, keyframe selection (KFS), also selects frames from video, yet
% focuses primarily on content and faces limitations in its evaluation
% protocols. 
% \yumo{Based on the task naming, one may wonder if hero frames are a subset of key frames: is "a frame being an key frame" an essential condition for "a frame being a hero frame"? We may want to clarify it somewhere. This affects whether we want the readers to see hero frame selection as an entirely new tasks (which deserve its own eval since it's so different from KFS), or an extension of KFS (where some eval can be reused, with some adaptation required).}
Compared with hero frame selection, which prioritizes visual impact, one related work, keyframe selection (KFS), selects frames rather for their information coverage. While the two tasks share a certain overlap in content relevance, existing KFS evaluation protocols exhibit limitations even on this front.
Annotation-based KFS benchmarks compare selected frames with reference
summaries, title-informed shot-importance scores, or question-relevant scenes
~\citep{gygli2014summe,song2015tvsum,li2026kfsbench}. These finite references
reduce content relevance to annotation agreement and cannot distinguish an
inferior selection from a valid alternative absent from the annotations.
VideoQA-based KFS benchmarks instead use the answer accuracy of a fixed VLM to
evaluate the selected frames
~\citep{mangalam2023egoschema,fu2025videomme,zhou2025mlvu,
wu2024longvideobench}. This protocol entangles frame-selection quality with
downstream VLM capability, since weaker VLMs may fail to extract the evidence
in informative frames, whereas stronger VLMs may answer correctly despite
redundant or even misleading frames. Moreover, KFS benchmarks largely draw on
user-generated, web, and egocentric videos rather than a dedicated movie
corpus~\citep{youtube2026about,grauman2022ego4d}. Most fundamentally,
their content-centered metrics insufficiently capture the aesthetic appeal
required of a movie's principal image. We therefore introduce
\textbf{\textsc{HeroFrame-Bench}}, the first benchmark for hero frame
movie's principal image through the joint assessment of movie-specific content
relevance, aesthetic appeal, and their interaction.

\begin{figure}[t]
    \centering
    \includegraphics[width=0.93\linewidth]{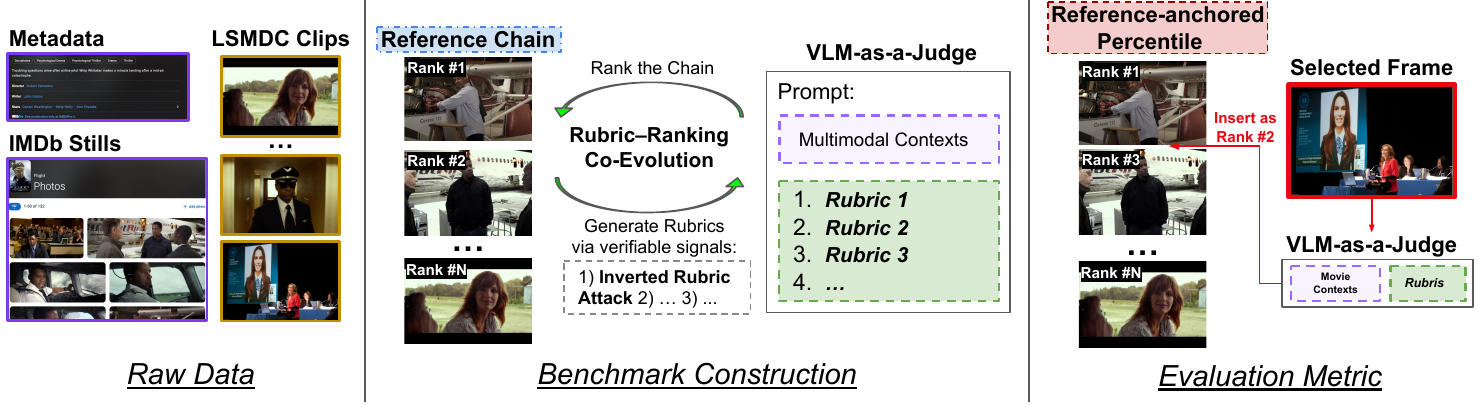}
    \vspace{-10pt}
    \caption{Overview of the \textsc{HeroFrame-Bench} construction and evaluation metric.}
    \label{fig:heroframe_overview}
    \vspace{-15pt}
\end{figure}

Constructing such an evaluator requires both broad semantic knowledge of the
movie and holistic visual judgment. We introduce VLM-as-a-Judge to direct
frame-selection evaluation, drawing on image generation domains
that use VLMs to assess similarly subjective semantic and perceptual qualities
~\citep{ku2024viescore,liu2025step1x}. 
We construct a movie corpus from LSMDC clips and captions
~\citep{rohrbach2017movie}, IMDb stills~\citep{imdb2026images}, and complementary
metadata from IMDb, TMDB, and MovieBench
~\citep{imdb2026titles,tmdb2026api,wu2025moviebench}
(Figure~\ref{fig:heroframe_overview}, left). We compact these sources
into multimodal contexts that situate the judgment within the movie's narrative
and visual identity.
As our evaluation metric, we propose the \emph{Reference-anchored Percentile}
(Figure~\ref{fig:heroframe_overview}, right), which is an insertion metric that calculates a selected frame's percentile
within pre-ranked reference chains formed by selecting IMDb stills and in-film
frames for broad coverage and visual diversity.
By using a VLM judge to automate ranking
through pairwise comparisons, our metric captures subjective preferences more
reliably~\citep{feng2026vab,narimanzadeh2023pairwise} than the pointwise metrics
used in generation fields~\citep{huang2023t2icompbench,ghosh2023geneval}.
Its pre-ranked reference chains also make our benchmark more readily extensible than some
Elo-ranked leaderboards
~\citep{chiang2024chatbotarena,jiang2024genaiarena} that require new human
judgments for each emerging method.
To ensure that the VLM judge produces reliable rankings, we further condition
it on explicit rubrics that reduce ambiguity and improve consistency in
subjective comparisons~\citep{kim2024prometheus,hashemi2024llmrubric}.
However, generating effective rubrics requires verifiable signals, such as 
the human preference labels or observed task outcomes used in existing
studies~\citep{liu2026openrubrics,ma2026skillgen,wu2026seed}, which may
not be readily available for hero frame selection.
We therefore develop
\emph{Rubric--Ranking Co-Evolution}
(Figure~\ref{fig:heroframe_overview}, middle), a reciprocal induction process.
Specifically, refined rubrics yield more reliable rankings that balance content
relevance and aesthetic appeal. In turn, the robustness of the stabilized
rankings supplies an internal verifiable signal for further rubric refinement.
This signal consists of three necessary conditions, most notably our
\emph{Inverted Rubric Attack}, 
which requires the frame order to reverse when a rubric's preference is semantically inverted.

Our contributions are threefold. 
\textbf{(1)} We introduce \textbf{\textsc{HeroFrame-Bench}}, the first benchmark
for hero frame selection from movie content, jointly evaluating movie-specific
content relevance, aesthetic appeal, and their interaction.
\textbf{(2)} We develop a scalable \emph{VLM-as-a-Judge} framework for direct
frame-selection evaluation, grounded in a multimodal movie corpus that
integrates diverse metadata.
The \emph{Reference-anchored Percentile} evaluates any
selected frame by inserting it into reusable, pre-ranked reference chains,
while \emph{Rubric--Ranking Co-Evolution} jointly refines rankings and evaluation
rubrics using internally verifiable necessary conditions centered on our
\emph{Inverted Rubric Attack}.
\textbf{(3)} We instantiate \textsc{HeroFrame-Bench} over 204 movies with 2,031
reference chains and 1,970 learned rubrics, and build an annotation interface
for human-alignment studies. The resulting human labels are used to validate the construction choices. The complete framework
improves agreement with human raters from 77.56\% to 83.78\%. 
% We further evaluate a broad suite of baselines. 
% \yumo{Consider saying a few more words about the eval results, particularly if the benchmark remains challenging for frontier models/systems $\longrightarrow$ hero frame selection as an important video understanding task is yet to be addressed $\longrightarrow$ the task and eval suite will likely be valued by the community.}
We further evaluate a broad suite of baselines. Most selectors fall below human-selected IMDb stills, with the previous advanced keyframe-selection method trailing the stills by 0.459, while the stills reach only 0.722 out of 1. Hero frame selection thus remains largely unaddressed, leaving substantial headroom for future methods

%% file: sections/02_related_work.tex
\section{Related Work}
\vspace{-8pt}

\subsection{Evaluation with Keyframe Annotations}
\vspace{-8pt}

Several benchmarks evaluate selected frames against human annotations. SumMe
provides reference summaries for each video, while TVSum provides importance
scores for each shot~\citep{gygli2014summe,song2015tvsum}.
Both constrain the summary length and compute an F measure from temporal overlap
with the annotations. The Yahoo Screen benchmark treats thumbnails chosen by
professional editors as references and counts a prediction as correct when its
SIFTflow distance from a reference falls below a fixed threshold. Its aesthetic
model uses manually designed features including brightness, sharpness, texture,
and composition~\citep{song2016click}. KFS-Bench annotates scenes needed to
answer each question, then measures how precisely and evenly the sampled frames
cover those scenes~\citep{li2026kfsbench}. Across these benchmarks, content
relevance is reduced to annotation overlap or a fixed visual distance, while
aesthetic quality is represented by manually specified image statistics such as
brightness and sharpness. These heuristic definitions are specific to
individual datasets and do not generalize to the joint assessment of content
and aesthetics. We instead use a VLM judge to assess both dimensions,
leveraging its broad knowledge and alignment with human preferences.

\subsection{Evaluation through Video Question Answering}
\vspace{-8pt}

Another evaluation protocol uses downstream VideoQA instead of keyframe
annotations. A fixed VLM answers questions using only the selected frames, and
its answer accuracy is treated as the score for keyframe selection. Long video
QA benchmarks suitable for this protocol include EgoSchema
~\citep{mangalam2023egoschema}, Video-MME~\citep{fu2025videomme}, MLVU
~\citep{zhou2025mlvu}, and LongVideoBench~\citep{wu2024longvideobench}. These
datasets annotate questions and answers for downstream video understanding
rather than which frames should be selected. The central limitation is the
entanglement between keyframe selection and the reasoning ability of the
downstream VLM. A weak VLM may fail despite sufficient evidence, while a strong
VLM may tolerate redundant or poorly selected frames and recover the answer from
partial visual evidence or language priors. VideoQA accuracy therefore cannot
isolate selection quality. It measures only content relevant to the annotated
questions and leaves aesthetic quality unmeasured. We instead
evaluate selected frames directly with a VLM judge over both content and
aesthetic criteria, removing the intervening QA task.

\subsection{VLM Judges and Rubric Generation}
\vspace{-8pt}

Since a video can contain multiple representative and aesthetically compelling
frames, no unique ground truth specifies the best selection. Similar ambiguity
arises in image generation and editing, where VLM judges evaluate semantic
fidelity and perceptual quality. VIEScore produces scores and rationales for
these dimensions~\citep{ku2024viescore}. GEdit-Bench applies VLM judges to
realistic editing tasks~\citep{liu2025step1x}. This subjectivity makes pairwise comparisons more
reliable than pointwise scores~\citep{feng2026vab} and less sensitive to
annotator disagreement and bias~\citep{narimanzadeh2023pairwise}. Image
generation arenas consequently aggregate pairwise votes by human into Elo rankings
~\citep{jiang2024genaiarena}. However, continuously collecting human pairwise judgments on newly emerging methods imposes substantial maintenance costs and hinders timely evaluation of new results. 
Such subjective pairwise evaluations are also common in LLM alignment~\citep{kim2024prometheus,hashemi2024llmrubric}, 
where VLM / LLM judges are leveraged to automate these comparisons across different rollouts, 
using explicit rubrics in judge prompts to minimize ambiguity and enhance consistency.
Yet effective rubric generation requires a verifiable signal for validating
candidate criteria. Reward modeling provides such a signal by preference labels
~\citep{christiano2017preferences,ouyang2022instructgpt}, which OpenRubrics uses
to filter criteria~\citep{liu2026openrubrics}. Skill
generation similarly uses task success or failure to validate generated skills.
SkillGen learns from successful and failed trajectories~\citep{ma2026skillgen}, while SEED distills trajectories labeled by
task success or failure~\citep{wu2026seed}. Keyframe selection, however, lacks a
comparably clear and verifiable signal. In this paper, we use a VLM
judge to obtain pairwise rankings anchored by reference frames, combining the
reliability of comparative evaluation with scalable automation. We then jointly
evolves rankings and rubrics so that each validates the other, supplying an
internal signal for rubric generation.

%% file: sections/03_method.tex
\section{Benchmark Construction}
\label{sec:construction}
\vspace{-8pt}

Each selection method takes only the movie as input (frames sampled at 1 FPS from its clips) and outputs selected frames. Given a selected frame, the VLM judge inserts it into the movie's frozen reference chains $\mathcal{C}$ under the frozen rubric $\mathcal{R}$, scoring the frame by its percentile position within $\mathcal{C}$ (Figure~\ref{fig:heroframe_overview}, right). Specifically, for each movie, we combine LSMDC clips and captions~\citep{rohrbach2017movie}, IMDb stills~\citep{imdb2026images}, and metadata from IMDb, TMDB, and MovieBench~\citep{imdb2026titles,tmdb2026api,wu2025moviebench}. The construction comprises three stages. First, to equip the VLM judge with the movie's narrative and visual context, the metadata are converted into a multimodal context, and diverse IMDb stills and in-film frames are assembled into reference groups (Figure~\ref{fig:heroframe_context}, Section~\ref{sec:context-groups}). Second, to make the VLM judge more reliable, Rubric--Ranking Co-Evolution ranks each group through pairwise VLM judgments and Bradley--Terry aggregation, then alternates rubric induction with group re-ranking, updating the movie-specific rubric $\mathcal{R}$ and the image rankings that define the reference chains $\mathcal{C}$ (Figure~\ref{fig:heroframe_build}, Section~\ref{sec:co-evolution}). Third, the frozen rubric and chains instantiate our metric, the Reference-Anchored Percentile (Figure~\ref{fig:heroframe_overview}, right, Section~\ref{sec:metric}).

\subsection{Multimodal Context and Reference Groups}
\label{sec:context-groups}
\vspace{-8pt}

\begin{figure}[t]
    \centering
    \includegraphics[width=0.95\linewidth]{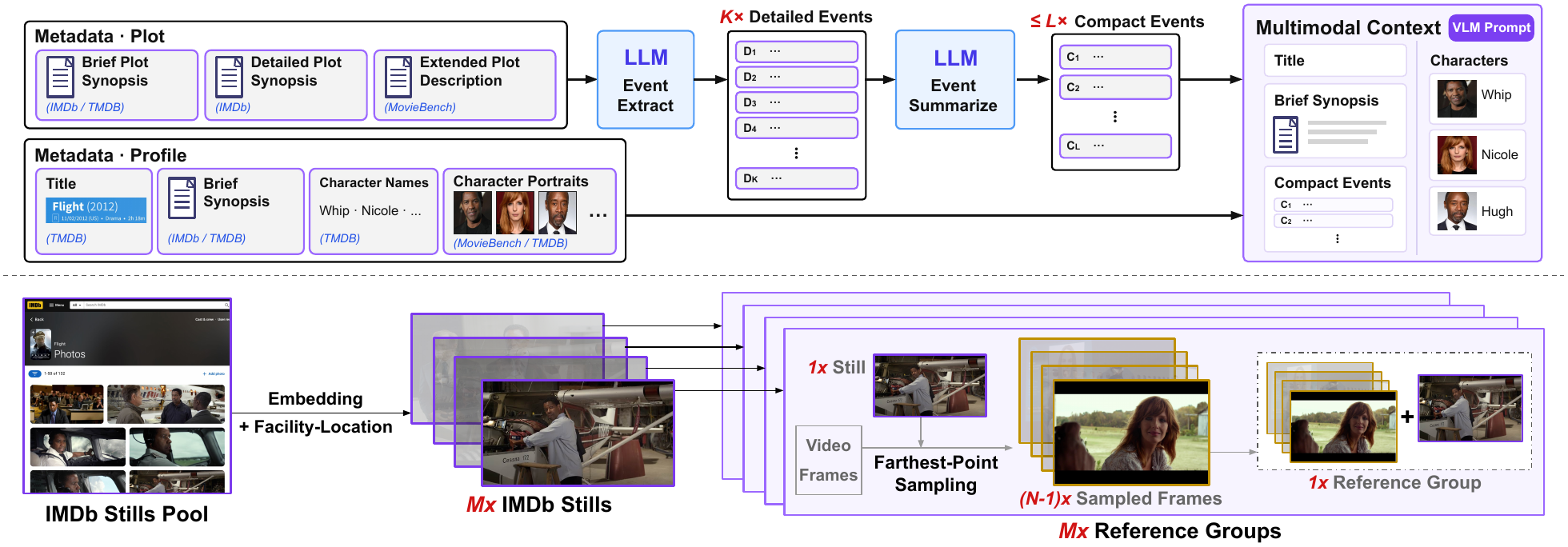}
    \vspace{-10pt}
    \caption{Multimodal context and reference groups. \emph{Top:}
    Detailed events are extracted from plot descriptions and summarized into
    compact events, which are combined with the movie profile to form the VLM
    context. \emph{Bottom:} Facility-location selects diverse IMDb
    stills, and each still is combined with diverse in-film frames selected by
    farthest-point sampling to form a reference group.}
    \label{fig:heroframe_context}
    \vspace{-10pt}
\end{figure}

\paragraph{Multimodal context.}
Figure~\ref{fig:heroframe_context} (top) presents the context construction.
We use GPT-5.6-sol for both LLM calls, with the model configuration specified
in Appendix~\ref{app:implementation}. We first prompt the LLM with the brief,
detailed, and extended plot descriptions and their source identifiers. The
LLM extracts a chronologically ordered list of the distinct plot events stated
in these sources. Each event contains a concise summary, explicitly named
characters, and supporting source references. We then prompt the LLM with the
ordered detailed events and a target output length. The LLM merges consecutive
events into compact narrative beats while preserving complete event coverage
and temporal order. The compact events are combined with the title, brief
synopsis, character names, and character portraits to form the multimodal
context used in every construction-stage VLM call. The complete extraction
and summarization prompts appear in
Appendix~\ref{app:prompts}, Figures~\ref{fig:prompt-event-extract}
and~\ref{fig:prompt-event-summarize}.

\vspace{-8pt}

\paragraph{Reference groups.}
Figure~\ref{fig:heroframe_context} (bottom) presents reference-group
construction. 
% We encode the valid IMDb stills with SigLIP2~\citep{siglip2},
% collapse near-duplicate images, and apply facility-location to select visually
% representative IMDb stills as anchors. 
We encode the valid IMDb stills with SigLIP2~\citep{siglip2}, collapse near-duplicate images, and apply facility-location~\citep{wei2015submodularity} to select visually representative IMDb stills as anchors, i.e., a compact core set in which every still is visually close to at least one selected still.
% \yumo{nit: consider citing for facility-location or adding a few more words on how it works. Folks outside of the data selection/curation community may not be familiar with the idea.} 
In parallel, we uniformly sample in-film
frames, remove visually uninformative and near-duplicate frames, and exclude
frames that duplicate an IMDb still. From this filtered pool, anchor-seeded
farthest-point sampling draws multiple visually diverse and mutually disjoint
companion sets for each anchor. Combining an anchor with one companion set
produces a candidate reference group. The filtering thresholds, embedding
model, facility-location objective, and sampling budgets are given in
Appendix~\ref{app:implementation}.

\subsection{Rubric--Ranking Co-Evolution}
\label{sec:co-evolution}
\vspace{-8pt}

\begin{figure}[t]
    \centering
    \includegraphics[width=0.95\linewidth]{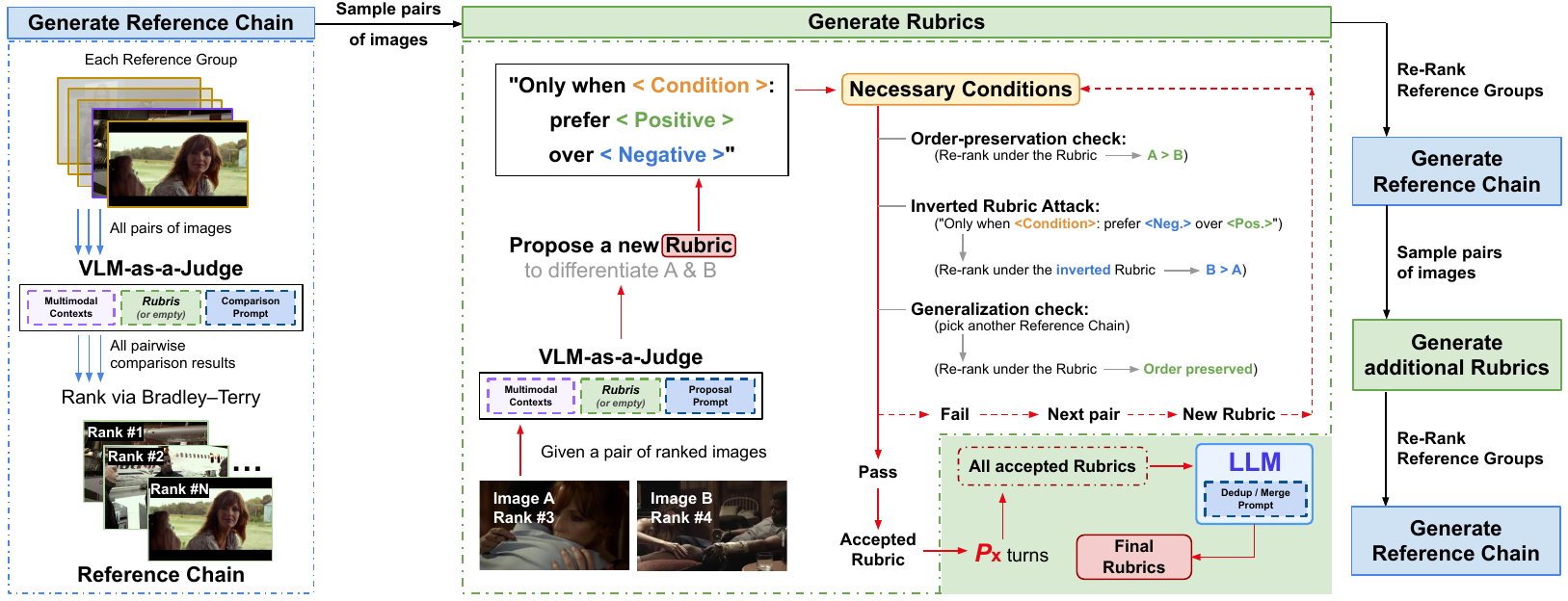}
    \vspace{-10pt}
    \caption{Rubric--Ranking Co-Evolution. Pairwise VLM judgments and
    Bradley--Terry aggregation rank each reference group into a reference chain
    (left). Ranked image pairs are sampled from reference chains
    for rubric generation. Each proposed rubric undergoes the
    Order-preservation check, Inverted Rubric Attack, and Generalization check
    before de-duplication (middle). The resulting rubrics re-rank the same
    groups. The updated reference chains generate another round of rubrics.
    With the completed rubrics held fixed, the final re-ranking forms the frozen
    reference chains (right).}
    \vspace{-10pt}
    \label{fig:heroframe_build}
\end{figure}

\paragraph{Generate reference chains.}
Each co-evolution round begins with the ranking procedure on the left of
Figure~\ref{fig:heroframe_build}. For every candidate reference group, the
GPT-5.6-sol receives each unordered image pair, the
multimodal context, the current rubric (empty for the initial ranking), and the
comparison prompt. We present each pair twice with its image placements
exchanged, and the VLM selects one winner in each call. The pair is decisive if
the VLM selects the same image twice, unresolved if it selects different
winners, and under-compared if either call abstains. Every non-abstaining call
contributes its resulting winner--loser ordering to a Bradley--Terry
model~\citep{bradley1952rank}:
\begin{equation}
    \Pr(i\succ j)=
    \frac{\exp(\theta_i)}{\exp(\theta_i)+\exp(\theta_j)},
    \label{eq:bt}
\end{equation}
where $\theta_i$ is the fitted strength of image $i$. Sorting the strengths
ranks the images within each candidate group. 
% A candidate group is eligible
% only if its anchor, the IMDb still, ranks in the upper half. 
% This rule favors the IMDb still, which is selected by humans, and treats a lower-half placement
% as evidence of an unreliable VLM judgment. 
A candidate group is eligible only if its anchor, the IMDb still, ranks in the upper half. Since the still is human-selected, this rule aligns the judge with human preference and treats a lower-half placement as evidence of an unreliable judgment. Section~\ref{sec:main-results} bears out this expectation, with IMDb stills scoring the highest among all tested selectors.
Among the eligible groups sharing
an anchor, we retain the group with the largest number of decisive pairs, using
fewer unresolved pairs and greater visual coverage as tie-breakers. The
retained group's image ranking becomes the reference chain for that anchor.
The ranked chain contains both high- and low-quality images, and the low-quality end is what lets the scale evaluate poor selected frames (Figure~\ref{fig:benchmark-example}).
Judge settings are specified in
Appendix~\ref{app:implementation}, and the comparison prompt appears in
Appendix~\ref{app:prompts}, Figure~\ref{fig:prompt-rank}.
\vspace{-8pt}

\paragraph{Generate rubrics from reference chains.}
For each reference chain, we select the \emph{most-tied} pair with the smallest
absolute difference between its fitted Bradley--Terry strengths and the
\emph{least-tied} pair with the largest difference. 
For each selected pair, the
judge receives the pair, the multimodal context, and the current rubric. An
additional proposal prompt withholds the reference-chain order and states that
the current rubric cannot separate the two images, presenting the pair as tied
despite its ranked origin. The prompt then requires the judge to commit to the
better hero frame and identify one uncovered visible distinction. 
The most-tied pair targets subtle distinctions the judge may have missed, while the least-tied pair turns obvious preferences into explicit rubrics, so that similar comparisons follow stated criteria rather than the judge's internal reasoning.
If no such
distinction exists, no rubric is proposed. Otherwise, the judge expresses the
distinction as one new rubric using the conditional preference form in the
middle of Figure~\ref{fig:heroframe_build}, which uses an \emph{Only when}
clause to specify the applicability condition and a
\emph{prefer ... over ...} relation to encode the directional preference.
Therefore, a preference
abstracted from a specific image pair applies only to pairs satisfying the stated
visible condition, limiting pair-specific overgeneralization.
Figure~\ref{fig:benchmark-example} shows a concrete case of a rubric proposed from such a tied pair.
Each anchor yields at most one accepted rubric from each pair type per
round. The rubric-generation constraints and retry budget are specified in
Appendix~\ref{app:implementation}, and the complete prompt appears in
Appendix~\ref{app:prompts}, Figure~\ref{fig:prompt-propose}.
\vspace{-8pt}

\paragraph{Verify Rubrics with Necessary Conditions.}
Each proposed rubric first passes a deterministic screen requiring a general,
visually grounded applicability condition, then undergoes the three
necessary-condition checks in the middle of
Figure~\ref{fig:heroframe_build}. First, the \emph{Order-preservation check}
requires the judge to select the same winner chosen during rubric generation
when re-ranking the source pair after adding it to the current rubric.
Second, our \emph{Inverted Rubric Attack} tests whether the judge
actually conditions on the proposed rubric. We exchange its Positive and
Negative terms while keeping its Condition unchanged, then re-rank the source
pair. Passing requires the winner to switch. Since the images and all other
context remain unchanged, this switch provides behavioral evidence that the
judge follows the rubric. Failure to switch indicates that the judge ignores
the supplied rubric and follows its internal prior. The attack therefore
filters rubrics that are linguistically plausible but behaviorally inactive.
Third, under the \emph{Generalization check}, the judge re-ranks the least
visually similar reference chain from another anchor after adding it to the
current rubric. The previously decisive relations in that reference chain must
remain largely stable. If a proposed rubric fails any check, the judge advances
to the next eligible image pair, generates a new rubric, and repeats all three
necessary-condition checks. Only rubrics that pass every check join the
accepted set.
Appendix~\ref{app:implementation} specifies the screens, acceptance scores,
cross-chain selection, and tolerance.
\vspace{-8pt}

\paragraph{De-duplicate Rubrics.}
GPT-5.6-sol proposes drops and merges among the accepted rubrics. A drop or
merge is applied only if the resulting rubric preserves every affected
source-pair winner and reverses each winner under inversion. The prompt is provided in Appendix~\ref{app:prompts},
Figure~\ref{fig:prompt-dedup}.
\vspace{-8pt}

\paragraph{Co-evolve and freeze.}
We run the preceding ranking and rubric-generation steps twice. Round one
starts from $\mathcal{R}_0=\varnothing$ and produces $\mathcal{R}_1$. Round two
re-ranks the same candidate groups under $\mathcal{R}_1$, reselects one
reference group per anchor, and generates additional rubrics from the updated
reference chains while conditioning on $\mathcal{R}_1$. We de-duplicate only
the new rubrics, then combine them with $\mathcal{R}_1$ to form
$\mathcal{R}_{\mathrm{final}}$. With $\mathcal{R}_{\mathrm{final}}$ fixed, the
judge re-ranks the reference groups selected in round two. We remove anchors
ranked in the lower half,
backfill deleted anchors when possible, and freeze the retained and replacement
reference chains as $\mathcal{C}_{\mathrm{final}}$. Evaluation uses
$(\mathcal{R}_{\mathrm{final}},\mathcal{C}_{\mathrm{final}})$. Deletion,
backfill, and de-duplication details appear in
Appendix~\ref{app:implementation}.
\vspace{-8pt}

\subsection{Reference-anchored Percentile}
\label{sec:metric}
\vspace{-8pt}

Each movie is associated with movie-specific final rubrics and a set of frozen
reference chains. We propose the \emph{Reference-anchored Percentile} to score a
selected frame $F$ against these chains using a VLM judge
conditioned on the chain, the rubrics, and a compact textual movie context. As shown in
Figure~\ref{fig:heroframe_overview} (right), for an $N$-image reference chain
$C=(c_1,\ldots,c_N)$ ordered from best to worst, the first call presents
$(c_1,\ldots,c_N)$ and returns $p_{\mathrm{fwd}}(F,C)\in
\{1,\ldots,N+1\}$, where position $1$ places $F$ above $c_1$ and position
$N+1$ places it below $c_N$. We then change the displayed order to
$(c_N,\ldots,c_1)$ and ask the judge to place $F$ again. The second prediction
$p_{\mathrm{rev}}(F,C)$ uses this new order, so position $1$ places $F$ below
$c_N$ and position $N+1$ places it above $c_1$. To improve robustness to
presentation order, we average the two estimates and normalize the result:
\begin{equation}
    \bar{p}(F,C)=
    \frac{p_{\mathrm{fwd}}(F,C)+N+2-p_{\mathrm{rev}}(F,C)}{2},
    \qquad
    s(F,C)=\frac{N+1-\bar{p}(F,C)}{N}\in[0,1].
    \label{eq:percentile}
\end{equation}
The resulting $s(F,C)$ is the fraction of reference images ranked below $F$.
Let $\mathcal{M}$ denote the benchmark movies, $\mathcal{F}_m$ the frames
returned for movie $m$, and $\mathcal{C}_m$ its frozen reference chains. We
average over reference chains for each frame, returned frames for each movie,
and movies:
\begin{equation}
    S=\frac{1}{|\mathcal{M}|}\sum_{m\in\mathcal{M}}
    \frac{1}{|\mathcal{F}_m|}\sum_{F\in\mathcal{F}_m}
    \frac{1}{|\mathcal{C}_m|}\sum_{C\in\mathcal{C}_m}s(F,C).
    \label{eq:benchmark-score}
\end{equation}
The VLM configuration is in
Appendix~\ref{app:implementation}, and the insertion prompt appears in
Appendix~\ref{app:prompts},
Figure~\ref{fig:prompt-insert}.

%% file: sections/04_experiments.tex
\section{Experiments}
\label{sec:experiments}
% \vspace{-8pt}

\begin{wraptable}{r}{0.49\linewidth}
\centering
\vspace{-74pt}
\caption{Dataset and construction statistics for \textsc{HeroFrame-Bench}.}
\label{tab:benchmark-statistics}
\footnotesize
\setlength{\tabcolsep}{3.5pt}
\renewcommand{\arraystretch}{0.98}
\begin{tabular}{@{}p{0.62\linewidth}>{\raggedleft\arraybackslash}p{0.32\linewidth}@{}}
\toprule
Statistic & Value \\
\midrule
Movies & 204 \\
Genres & 18 \\
Release years & 1943--2014 \\
LSMDC clips & 128,085 \\
Total clip duration & 140.8 h \\
Median duration per movie & 39.5 min \\
Reference chains & 2,031 \\
Images per reference chain & 6 \\
IMDb stills & 2,031 \\
Learned rubrics & 1,970 \\
Rubrics / movie (min./med./max.) & 2 / 10 / 18 \\
IMDb still win rate & 82.22\% \\
Human select still top-3 rate & 80.0\% \\
\bottomrule
\end{tabular}
\vspace{-12pt}
\end{wraptable}
\subsection{Benchmark Statistics}
\label{sec:benchmark-statistics}

Table~\ref{tab:benchmark-statistics} summarizes the scale and composition of
\textsc{HeroFrame-Bench}. It covers 204 films across 18 genres. 
The 128,085 LSMDC clips provide 140.8 hours of video,
with a median duration of 39.5 minutes per film after concatenation. Each of the 2,031 frozen
reference chains contains one IMDb still and five sampled frames. The IMDb still 
wins 82.22\% of comparisons in our construction, and humans place it in the top three for 80.0\% of
groups in the following human study. The benchmark contains 1,970 learned rubrics, 
with a median of ten per
film and a range of two to eighteen. 
Each rubric contains 27--54 words, with a median of 36, and the collection
spans a vocabulary of 1,704 words.
During rubric generation, the VLM also produces a short descriptive name to
label each rubric, such as \emph{Subject legibility}, \emph{Facial visibility},
and \emph{Dramatic tension}. The 1,970 rubrics comprise 666 distinct names. The most frequent
terms include
\emph{legibility}, \emph{subject}, \emph{facial}, \emph{visibility}, and
\emph{character}.

\subsection{Benchmark and Evaluation Configuration}
\label{sec:setup}

GPT-5.6-Sol constructs the multimodal context and serves as the construction
judge for reference-group ranking, rubric generation, necessary-condition
checks, de-duplication, and final re-ranking. For evaluation, we sort
each film's LSMDC clips~\citep{rohrbach2017movie} by start time, concatenate
them, and sample frames at 1 FPS. Each frame selection method receives only
these frames, with its output limited to five frames. It does not receive the multimodal
context, final rubric, IMDb stills, or frozen reference chains. The benchmark
adopts Qwen3.6-27B~\citep{qwen36} to 
score each selected frame, avoiding recurring API costs during evaluation. 
The model receives
the selected frame together with the title, year, brief synopsis, final rubric, and frozen
reference chain. Appendices~\ref{app:implementation} and~\ref{app:prompts}
give the complete model configurations and prompts.

\subsection{Human Study and Ablations}
\label{sec:ablations}

\begin{table}[h]
\centering
\vspace{-16pt}
\caption{Human annotation interface and benchmark construction ablations.}
\begin{minipage}[t]{0.505\linewidth}
  \vspace{0pt}
  \centering
  \textbf{(a) Annotation interface}\par\vspace{3pt}
  \includegraphics[width=\linewidth]{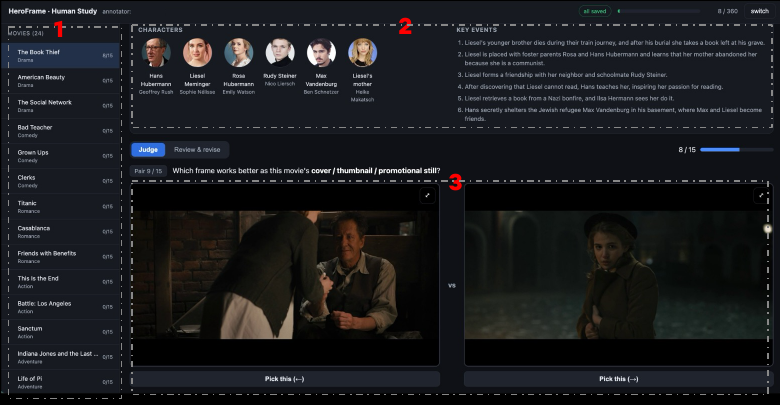}
\end{minipage}\hfill
\begin{minipage}[t]{0.475\linewidth}
  \vspace{0pt}
  \centering
  \textbf{(b) Construction ablation}\par\vspace{2pt}
  \scriptsize
  \setlength{\tabcolsep}{2.3pt}
  \renewcommand{\arraystretch}{1.08}
  \begin{tabular}{@{}c p{0.55\linewidth}rr@{}}
    \toprule
    ID & Construction & \makecell{Avg. \#\\rubrics} & \makecell{Agree.\\(\%)} \\
    \midrule
    a) & GPT-5.6-Sol & 0.00 & 77.56 \\
    b) & \makecell[l]{Most-tied pair +\\necessary-condition checks} & 2.10 & 80.67 \\
    \rowcolor{herogray}
    c) & b) $-$ necessary-condition checks & 2.10 & 78.44 \\
    \rowcolor{herogray}
    d) & b) + hand-written rubrics & 5.20 & 79.33 \\
    e) & b) + least-tied pair & 9.37 & 81.78 \\
    f) & e) + de-duplication & 8.23 & 82.44 \\
    \rowcolor{heropurple!11}
    \textbf{g)} & \textbf{f) + second round (ours)} & \textbf{9.87} & \textbf{83.78} \\
    h) & g) + third round & 11.07 & 82.44 \\
    \addlinespace[2pt]
    \rowcolor{herocyan!10}
    i) & g), GPT-5.6-Sol $\rightarrow$ Qwen3.6-27B & 10.83 & \textbf{82.00} \\
    \bottomrule
  \end{tabular}
\end{minipage}
\label{tab:human-study}
\vspace{-12pt}
\end{table}

Table~\ref{tab:human-study}(a) shows the interface used to obtain pairwise human
preferences. Its left panel (labeled as red 1) lets annotators move among movies. The upper context
panel (labeled as red 2) presents character names and portraits together with key narrative
events. The central comparison panel (labeled as red 3) shows a randomized frame pair and asks
which image would better serve as the movie's promotional
entry point. Arrow-key input supports rapid comparison while the multimodal context
remains visible.
We sample three films from each of the ten most frequent TMDB genres, with no
repeated film. 
For each film, one six-image group contributes all 15 unordered pairs, yielding
30 groups and 450 human annotations. 
For each labeled pair, the VLM judge evaluates both image orders. We report
agreement as the percentage of pairs for which both selections
match the human preference. Appendix~\ref{app:human-study} gives the full protocol.

Table~\ref{tab:human-study}(b) shows that, without any rubric, GPT-5.6-Sol achieves 77.56\% agreement, indicating that a
VLM judge can already align reasonably well with human preferences through its
training. Our \emph{Rubric--Ranking Co-Evolution} further raises agreement to 83.78\%.
During construction, adding rubrics from the most-tied pair together with the
necessary-condition checks yields a +3.11\% improvement. Removing
these checks lowers agreement from 80.67\% to 78.44\%, showing that learned
rubrics are most effective when unreliable proposals are filtered. Adding
human-written rubrics instead causes a slight drop, suggesting that generic
human priors are less suitable than movie-specific induction. Drawing rubrics
from the least-tied pair and applying de-duplication further raises agreement to
82.44\%. A second co-evolution round peaks at 83.78\%, whereas a third round
provides no further gain while increasing the burden of eliciting rubrics that
differ from those already learned. Replacing GPT-5.6-Sol with open-weight
Qwen3.6-27B still retains 82.00\% agreement, +4.44\% above
GPT-5.6-Sol without learned rubrics. This result shows that the construction of
\textsc{HeroFrame-Bench} does not depend heavily on a strong VLM and provides a
scalable, transferable framework.

\subsection{Evaluation Analysis and Results}
\label{sec:main-results}

Although the \emph{Reference-anchored Percentile} and
\emph{Rubric--Ranking Co-Evolution} use nearly identical judging prompts,
construction derives each chain order from pairwise comparisons, whereas
evaluation directly inserts a candidate into the frozen chain. 
As shown in Table~\ref{tab:evaluation-analysis},
we therefore test \emph{ranking fidelity}, whether direct
insertion preserves the construction's pairwise order. For each chain, we hold
out one image and recover its position by insertion into the five-image
remainder or by five pairwise comparisons against the remaining images. 
% \yumo{Five pairwise comparisons are needed while Table 3 lists 6x eval calls - did I miss something?}
\begin{wraptable}{r}{0.47\linewidth}
\centering
\vspace{-15pt}
\caption{Evaluation and construction analysis.}
\label{tab:evaluation-analysis}
\vspace{3pt}
\footnotesize
\setlength{\tabcolsep}{2pt}
\renewcommand{\arraystretch}{0.98}
\begin{tabular}{@{}p{0.52\linewidth}rr@{}}
\toprule
\rowcolor{heropurple!8}
\multicolumn{3}{@{}l}{\textbf{(a) Ranking fidelity}} \\
Evaluation form & \makecell{Spearman rank\\correlation ($\rho$)} & \makecell{Eval.\\calls} \\
\midrule
Pairwise comparison & 0.988 & $5\times$ \\
Direct insertion & 0.973 & $1\times$ \\
\addlinespace[1pt]
\rowcolor{heropurple!8}
\multicolumn{3}{@{}l}{\textbf{(b) Movie specificity}} \\
Rubric condition & \multicolumn{2}{r@{}}{Agree. (\%)} \\
\midrule
No rubric & \multicolumn{2}{r@{}}{75.56} \\
Mismatched rubric & \multicolumn{2}{r@{}}{78.00} \\
Movie's own rubric & \multicolumn{2}{r@{}}{\textbf{82.00}} \\
\addlinespace[1pt]
\rowcolor{heropurple!8}
\multicolumn{3}{@{}l}{\textbf{(c) Rubric admission}} \\
Round & \multicolumn{2}{r@{}}{Rejected proposals (\%)} \\
\midrule
First & \multicolumn{2}{r@{}}{33.4} \\
Second & \multicolumn{2}{r@{}}{94.4} \\
Third & \multicolumn{2}{r@{}}{94.9} \\
\bottomrule
\end{tabular}
\vspace{-10pt}
\end{wraptable}
We use Spearman rank
correlation ($\rho$), where $1$, $-1$, and $0$ denote perfect positive, perfect
negative, and no monotonic correlation, respectively. 
Direct insertion reaches
$\rho{=}0.973$, close to $0.988$ for pairwise recovery, with one fifth as many
evaluation calls. 
We next test \emph{movie specificity}: whether learned
rubrics encode source-movie criteria beyond generic hero-frame preferences.
For the \emph{mismatched rubric} condition, we permute the movie-level rubrics
so that every movie receives a rubric constructed from a different movie. Using Qwen3.6-27B
on the same 450 human-labeled pairs, agreement is 75.56\% without a rubric,
78.00\% with mismatched rubrics, and 82.00\% with each movie's own rubrics. The
improvement under mismatch indicates shared criteria across movies. The
\emph{Only when} clause explicitly limits each rubric to its applicability
conditions, so that it remains
inactive rather than reducing agreement. The further gain from each movie's
own rubric shows that most of the benefit is movie-specific. Finally,
\emph{rubric admission} tests whether later co-evolution rounds continue to
produce distinct rubrics that satisfy the necessary condition checks. Since each new
round must avoid criteria already contained in the current rubric, the
rejection rate rises from 33.4\% in the first round to 94.4\% and 94.9\% in the
next two.

\begin{wraptable}{r}{0.48\linewidth}
\centering
\vspace{-22pt}
\caption{Hero-frame selection results.}
\label{tab:keyframe-results}
\vspace{3pt}
\small
\setlength{\tabcolsep}{3pt}
\renewcommand{\arraystretch}{1.2}
\begin{tabular}{@{}p{0.76\linewidth}r@{}}
\toprule
Method / source & $S$ \\
\midrule
\rowcolor{herocyan!10}
IMDb still & 0.722 \\
Largest-face heuristic & 0.557 \\
\rowcolor{herogray}
Random sampling & 0.501 \\
Uniform sampling & 0.439 \\
SigLIP2 text retrieval & 0.362 \\
Brightness + sharpness & 0.297 \\
WFS-SB~\citep{wfssb} & 0.263 \\
\bottomrule
\end{tabular}
\vspace{-10pt}
\end{wraptable}
We then conduct evaluation on several baselines in Table~\ref{tab:keyframe-results}, 
including uniform sampling, largest-face and brightness-and-sharpness
heuristics, SigLIP2 retrieval, and WFS-SB~\citep{wfssb}, a prior advanced
open-source keyframe-selection method. IMDb stills and random frames serve as
human-selected and random references. IMDb stills score $0.722$, compared with
$0.501$ for random frames. Their advantage confirms that human selection
provides a useful quality prior. However, because IMDb stills can be uploaded
by general users without an explicit professional frame-selection protocol,
they do not always align with human hero-frame preferences. This observation
supports our construction choice to favor reference chains in which the IMDb
still ranks highly without requiring it to rank first. The largest-face
heuristic is the only tested selector above random, reaching $0.557$. Character
faces therefore provide a useful heuristic for hero-frame selection, but the
$0.165$ gap to IMDb stills shows that face size alone does not generalize to the
full task. Uniform sampling, SigLIP2 retrieval, brightness and sharpness, and
WFS-SB all fall below random. These results show that temporal coverage,
similarity to a hero-frame text query, fixed photographic cues, and
semantic-boundary-based segmentation each optimize an insufficient proxy for
the movie-specific relevance and visual appeal required of a hero frame.

\begin{figure}[t]
\centering
\includegraphics[width=\linewidth]{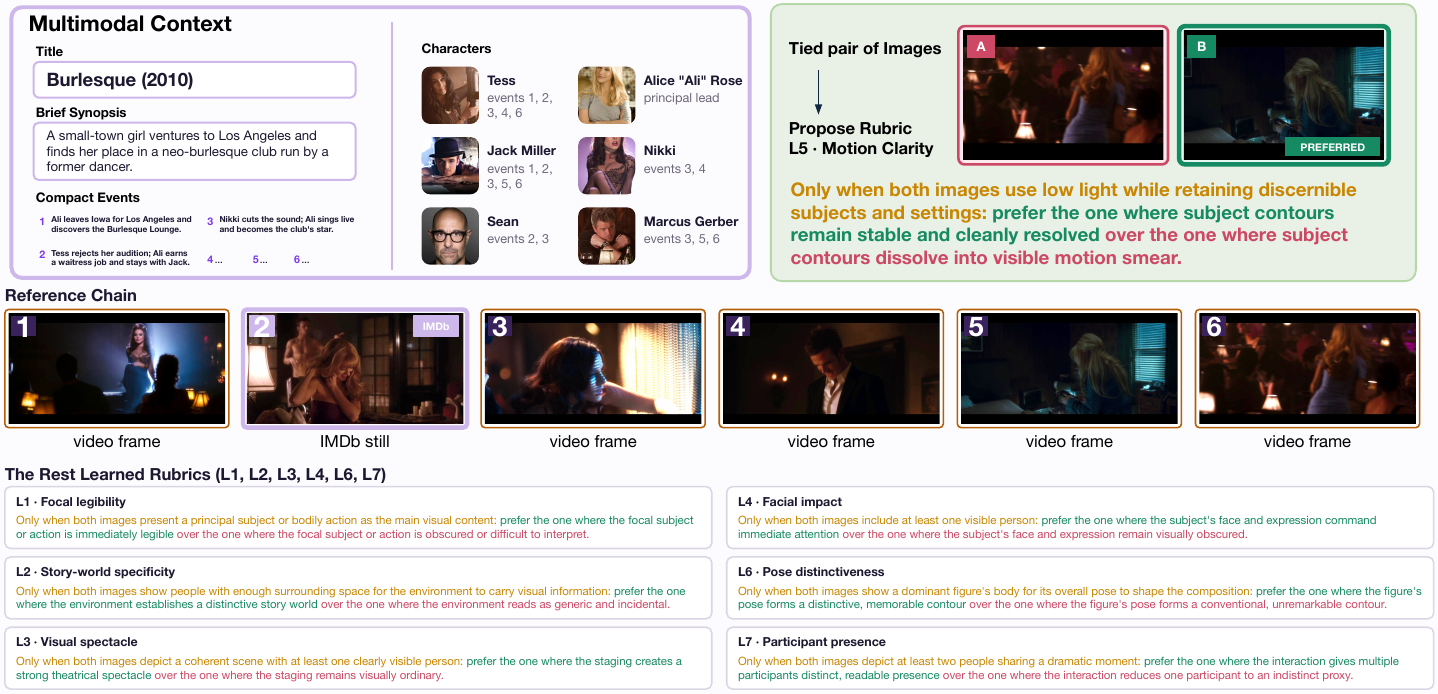}
\vspace{-16pt}
\caption{Construction example for \emph{Burlesque} (2010), showing the
multimodal context, a rubric proposal from a tied pair of images, the reference
chain, and the learned rubrics.}
\label{fig:benchmark-example}
\end{figure}
\vspace{-10pt}

\subsection{Qualitative result and Limitation}
\label{sec:further-analysis}

\paragraph{Visualization.} Figure~\ref{fig:benchmark-example} visualizes the
benchmark construction for \emph{Burlesque} (2010). The upper-left panel shows
the multimodal context, including the title, synopsis, compact narrative
events, and character portraits, with each character linked to the events in
which they appear. This evidence helps the VLM ground its judgments in a
complex and subjective task. The upper-right panel shows a rubric proposal
from a tied pair of images during \emph{Rubric--Ranking Co-Evolution}. Both
images are weak hero-frame candidates and later rank near the bottom of the
reference chain, yet the VLM must still order them. The proposed \emph{Motion
Clarity} rubric uses its \emph{Only when} clause to specify that both images
are low-light scenes with discernible subjects. Since the principal character
faces away from the camera in both images, facial evidence does not distinguish
them. The VLM instead identifies stable, resolved contours in the preferred
image and visible motion smear in the other. Including low-ranked, low-quality
images in the reference chain is necessary because they provide reference
points for evaluating poor selected frames. A chain containing only
high-quality images would lack this lower range of the evaluation scale. The
reference chain shown in the middle places the user-uploaded IMDb still second.
The still presents a recognizable narrative moment and the principal character,
but the principal character's facial expression is not clearly resolved. 
This example demonstrates that IMDb uploads can generally align with human, although the quality is not always consistently high due to the lack of explicit, professional guidelines for selection.
 In contrast, the
first-ranked frame is sampled directly from the video. Its balanced
golden-ratio composition, eye-catching stage lighting, and clear rendering of
the principal character's face and expression create strong visual appeal,
while the nightclub performance directly conveys the film's club-dancer theme.
The frame therefore achieves a stronger balance between aesthetics and
movie-specific content, demonstrating that video sampling can recover a better
still than the available IMDb still. The remaining rubrics extend the
comparison across face legibility, facial impact, story-world specificity,
participant presence, visual spectacle, and pose distinctiveness. Their
conditional forms restrict each rubric to relevant comparisons.

\paragraph{Limitations.} Copyright and data access currently restrict the
benchmark to LSMDC movies. Additional licensed data could broaden its coverage
of films, languages, and production settings. Current VLM context capacity also
limits each reference chain to six images. More capable VLMs could support
longer chains and a finer-grained scale, provided judgment fidelity is preserved.

\FloatBarrier
\flushbottom

%% file: sections/05_conclusion.tex
\section{Conclusion}
\vspace{-8pt}

We introduced \textsc{HeroFrame-Bench}, the first benchmark to jointly assess
movie-specific relevance and aesthetic appeal in hero-frame selection.
Its construction treats benchmark design as an iterative measurement problem.
\emph{Rubric--Ranking Co-Evolution} uses multimodal context and verifiable
necessary conditions, including the \emph{Inverted Rubric Attack}, to induce
movie-specific rubrics and stable reference chains. The
\emph{Reference-anchored Percentile} converts these chains into a reusable
evaluation scale. Across 204 movies, 2,031 chains and 1,970 learned rubrics
increase human agreement from 77.56\% to 83.78\%, while retaining 82.00\% with
open-weight Qwen3.6-27B. The best tested automatic selector scores only $0.557$
against $0.722$ for human-selected IMDb stills, and most baselines fall below
random sampling. This gap confirms the need for a stable, movie-conditioned
evaluation scale and more capable selectors that balance narrative relevance
with visual appeal.

%% file: sections/06_ai_use_statement.tex
\clearpage
\subsection*{AI use statement}

Generative AI tools assisted with language revision and LaTeX typesetting. The
authors determined the scientific content, verified all claims and reported
values against the underlying experiments, and take responsibility for the
final manuscript.

%% file: sections/07_ethics_statement.tex
\subsection*{Ethics statement}

The benchmark studies images from films and inherits biases in the LSMDC movie
collection, IMDb promotional stills, and the VLM judges. It should not be used
to infer sensitive attributes or make judgments about depicted individuals.
Human annotators compare frames for a narrowly defined promotional function;
the study does not solicit demographic or other sensitive personal data. To
respect source licensing, the public frame track is distributed as a
reproducible index into LSMDC clips rather than as redistributed movie video.

%% file: sections/08_reproducibility_statement.tex
\subsection*{Reproducibility statement}

Section~\ref{sec:construction} defines benchmark construction and
Section~\ref{sec:setup} defines the evaluation protocol. Appendix
~\ref{app:implementation} reports sampling, ranking, gating, and reader
settings; Appendix~\ref{app:prompts} provides prompt templates; and Appendix
~\ref{app:human-study} details the human study. The release contains the final
rubrics and their provenance, ordered chains and pairwise decisions, frozen
prompt versions, raw model outputs, a 1-fps track index, verification scripts,
and baseline implementations.

%% file: sections/09_appendix.tex
\section*{Appendix}

\section{Construction and Evaluation Details}
\label{app:implementation}

\paragraph{Multimodal context.}
The construction judge receives the title and year, one brief synopsis, at most
eight character names with reference portraits, and at most six compact key
events. Detailed events are extracted from IMDb, TMDB, and MovieBench plot
sources before a second pass compresses them. Source quotations and event
links remain in the audit record but are not shown to the judge. Final
evaluation uses the textual portion of the multimodal context, containing the
title, year, and one-sentence synopsis, together with the final rubric. It does
not include character portraits or key events. Both event-processing calls use
GPT-5.6-Sol. Event extraction and each hierarchical summarization call use
maximum completion budgets of 32,000 and 16,000 tokens, respectively.

\paragraph{Construction judge.}
All reference-group ranking, rubric generation, necessary-condition checks,
and de-duplication calls use GPT-5.6-Sol. Candidate images and character portraits are encoded with
longest edges of 768 and 640 pixels, respectively. Ranking and rubric-generation
calls use a maximum completion budget of 8,000 tokens. De-duplication calls use
16,000 tokens.

\paragraph{Reference pools.}
IMDb images must decode as RGB still frames and pass size checks.
SigLIP2~\citep{siglip2}
embeddings define connected components at cosine similarity $0.95$, from
which one image is retained before facility-location chooses ten anchors. The
LSMDC pool is sampled at 2.5-s bucket midpoints. Frames with grayscale standard
deviation below 12 are removed. It is de-duplicated internally and against the
IMDb still pool at cosine $0.95$, then farthest-point sampled to at most 500
frames. Each anchor receives up to four disjoint draws of five sampled frames.

\paragraph{Ranking and reference-group selection.}
Every six-image ranking uses 15 unordered pairs and two placements per pair.
Every non-abstaining directional decision contributes one ordering to the
Bradley--Terry fit. A pair is decisive only when both placements select the
same winner. A candidate group is eligible only when the anchor ranks in the
top three. Among eligible groups sharing an anchor, the group with the most
decisive pairs is retained, using fewer unresolved pairs and greater visual
coverage as tie-breakers. An anchor is deleted during final freeze if it ranks
fourth or below.

\paragraph{Rubric screening and necessary-condition checks.}
For each of the most-tied and least-tied pair types, the judge makes at most
three rubric-generation attempts.
After a proposed rubric fails, the next attempt advances to the next eligible
pair and may continue into another draw for the same anchor.
Before model-based validation, a deterministic screen rejects an empty
applicability condition, a condition longer than 20 words, character names,
color/wardrobe/time-of-day terms, or a condition that restates the preferred
end. In the Inverted Rubric Attack, source-pair support is scored as 1 for a
decisive winner, $0.5$ for a Bradley--Terry lead, and 0 for a loss. A proposed
rubric passes only when the source-pair winner receives positive support under
the proposed rubric, the opposite winner receives positive support under its
inverted form, and at least one direction is decisive. The Generalization
check rejects a proposed rubric if it reverses more than
$\max(1,\lfloor0.15n\rfloor)$ of the $n$ previously decisive pairs in the
least visually similar reference chain from another anchor, identified by
mean-pooled SigLIP2 cosine similarity.

\paragraph{Rounds and final freeze.}
Round one starts from an empty rubric. Round two starts from the de-duplicated
first-round rubric. Both rounds use the same candidate groups sampled before
ranking, but each round reselects one reference group from the updated
rankings. Round-two inverted attacks include the complete inherited rubric,
and de-duplication can modify only newly learned rubrics. Final freeze re-ranks
the reference group selected in round two under the combined rubric, prunes
rubrics originating from deleted anchors, and backfills deleted anchors when
possible.

\paragraph{Evaluation with Qwen3.6-27B.}
Methods choose at most five frames from candidate frames sampled at 1 FPS.
Qwen3.6-27B is served locally with candidate images encoded at 1024 pixels,
temperature $0.6$, thinking enabled, a maximum context length of 16,384 tokens,
and a maximum completion budget of 16,000 tokens. Prefix caching is disabled.
The model receives the title, year, one-sentence synopsis, final rubric, frozen
reference chain, and candidate frame. The insertion prompt is issued in both best-to-worst
and worst-to-best directions. If one response is invalid, the valid response is
used alone. The score is omitted only when both responses are invalid. Per-frame
scores average over all reference chains with resolved insertions for the movie
(2,031 chains in total, mean 9.96 per film). Text-conditioned frame-selection
methods receive exactly: ``Pick a hero frame from this movie. A hero frame is
a single still image taken from within a movie that would work as its cover,
thumbnail, or promotional entry point.''

\clearpage
\section{Full Prompt Templates}
\label{app:prompts}

The following templates preserve the functional text and placeholders of the
frozen prompts. The multimodal context and numbered rubric are inserted
verbatim from the released records. Within the prompt text, ``guideline'' is
the user-facing rendering of an individual rubric. Necessary-condition checks
use the ranking template with the proposed rubric in its natural or inverted
form.

\begin{figure}[!htbp]
\promptrole{System}
\begin{promptbox}
You compile source-grounded movie evidence. Use only supplied evidence.
Every conclusion must cite supplied source IDs. Return one valid JSON object,
with no markdown.
\end{promptbox}
\promptrole{User}
\begin{promptbox}
<MOVIE TITLE, ID, AND GENRES>

Extract all plot events supported by the supplied plot texts. Cover the full
storyline in narrative order, including the setup, intermediate events, and
ending. Split distinct events and do not add movie knowledge beyond the
sources. For each event, return a concise summary, explicitly named
characters, and at least one short supporting quote with its source ID.

Return only JSON:
{"canonical_events":[{"event_id":"E1","summary":"...",
"characters":["..."],"source_refs":[{"source_id":"...",
"quote":"..."}]}]}

<PLOT TEXTS WITH SOURCE IDS>
\end{promptbox}
\caption{Event-extraction prompt. The inputs contain the available brief,
detailed, and extended plot descriptions with provenance identifiers.}
\label{fig:prompt-event-extract}
\end{figure}

\begin{figure}[!htbp]
\promptrole{System}
\begin{promptbox}
You compile source-grounded movie evidence. Use only supplied evidence.
Every conclusion must cite supplied source IDs. Return one valid JSON object,
with no markdown.
\end{promptbox}
\promptrole{User}
\begin{promptbox}
<MOVIE TITLE>
<TARGET NUMBER OF OUTPUT EVENTS>

Merge the numbered story events into the target number of higher-level
events while preserving strict narrative order. Merge only consecutive
events. The output groups must cover every input index exactly once. Each
output summary must faithfully cover every event it absorbs and must not
introduce facts absent from the inputs.

Return only JSON:
{"events":[{"summary":"...","absorbs":[1,2,3]}, ...]}

<NUMBERED DETAILED EVENTS>
\end{promptbox}
\caption{Event-summarization prompt. Longer event sequences are merged
hierarchically by applying this template at successive levels.}
\label{fig:prompt-event-summarize}
\end{figure}

\begin{figure}[!htbp]
\promptrole{System}
\begin{promptbox}
You are an expert film editor. A hero frame is a single still image taken
from within a movie that would work as its cover, thumbnail, or promotional
entry point. You are shown two candidate stills from one movie and decide
which is the better hero frame, judging only by the images, the movie
information you are given, and the numbered guidelines. You are precise and
consistent.
\end{promptbox}
\promptrole{User}
\begin{promptbox}
<MOVIE CONTEXT>

GUIDELINES FOR JUDGING A HERO FRAME
Weigh them all together; when they point in different directions, none of
them automatically overrides the others.
A guideline beginning "Only when both images ..." applies only if BOTH
candidates meet that condition. Otherwise neither candidate is penalised.

<NUMBERED RUBRIC>

Decide which of A, B is the better hero frame and give one short reason. If
the two are genuinely too close to separate, leave the winner empty.
Return only JSON: {"winner":"<A, B, or empty>","reason":"<sentence>"}.
\end{promptbox}
\caption{Pairwise ranking prompt. With an empty rubric, the guideline block is
omitted while the task and output contract remain unchanged.}
\label{fig:prompt-rank}
\end{figure}

\begin{figure}[!htbp]
\promptrole{System}
\begin{promptbox}
You are an expert film editor who writes clear, reusable guidelines for
judging whether a still image works as a movie cover. A hero frame is a single
still image taken from within a movie that would work as its cover, thumbnail,
or promotional entry point. Judge only what appears in the image itself; each
guideline must apply to images from any movie, not one particular picture.
\end{promptbox}
\promptrole{User}
\begin{promptbox}
<MOVIE CONTEXT>
<CURRENT GUIDELINES, IF ANY>

The current guidelines cannot separate A and B. Commit to which is the better
hero frame; name the single visible, non-marginal quality that decided it and
is not already covered; give the general condition BOTH images must meet.
Write one rule as:
"Only when both images <condition>: prefer the one where <preferred end>
 over the one where <opposite end>."

The condition must be a general verb phrase under 20 words, must not name a
character, place, color, garment, or time of day, and must not restate the
preferred end. Return only JSON with better and one rule containing name,
when, prefer, and over. If no uncovered distinction exists, return no rule.
\end{promptbox}
\caption{Rubric-proposal prompt. Rejected rubrics are appended as a
do-not-repeat list on retries.}
\label{fig:prompt-propose}
\end{figure}

\begin{figure}[!htbp]
\promptrole{System}
\begin{promptbox}
You are reviewing guidelines that were each written separately. Identify
where the set repeats itself.
\end{promptbox}
\promptrole{User}
\begin{promptbox}
<NUMBERED RUBRIC>

Find guidelines that genuinely say the same thing. For each overlap, either
drop a guideline already covered by another or merge partial statements of
one quality into a single applicability-gated prefer/over rule. Each guideline
may appear in at most one operation. If every guideline says something the
others do not, change nothing. Return only JSON containing drop/merge ops.
\end{promptbox}
\caption{De-duplication prompt. Proposed operations are applied only after
validation on the source pairs of all affected rubrics.}
\label{fig:prompt-dedup}
\end{figure}

\begin{figure}[!htbp]
\promptrole{System}
\begin{promptbox}
You are an expert film editor. A hero frame is a single still image taken
from within a movie that would work as its cover, thumbnail, or promotional
entry point. You are shown stills already ranked from best to worst, plus one
candidate, and decide where the candidate belongs using the movie information
and numbered guidelines.
\end{promptbox}
\promptrole{User}
\begin{promptbox}
<TITLE, YEAR, ONE-SENTENCE SYNOPSIS>
<NUMBERED RUBRIC>

Images A--F are already ordered from BEST to WORST. Image G is not yet placed.
Return its position from 1 to 7 and one short reason: 1 is better than all six,
7 is worse than all six. Return only JSON with position and reason.
\end{promptbox}
\caption{Direct-insertion prompt. A second call reverses both the chain and
the textual direction, counts from the worst end, and is mapped back before
the two positions are averaged.}
\label{fig:prompt-insert}
\end{figure}

\FloatBarrier

% \section{Qualitative Benchmark Examples}
% \label{app:qualitative-examples}

% Figures~\ref{fig:example-halloween}--\ref{fig:example-district9} show three
% additional movies selected from the released benchmark. Every chain is fully decisive, ranks a video frame
% first, and retains its IMDb still at rank two. Each visualization also shows
% six learned rubrics spanning different aspects of hero-frame quality.

% \begin{figure}[!htbp]
% \centering
% \includegraphics[width=\linewidth]{figures/halloween_demo.pdf}
% \vspace{-4pt}
% \caption{Qualitative example from \emph{Halloween}. The top-ranked video
% frame supplies a stronger character-centered expression of threat than the
% IMDb still while preserving the still as a strong second-ranked reference.}
% \label{fig:example-halloween}
% \end{figure}

% \begin{figure}[!htbp]
% \centering
% \includegraphics[width=\linewidth]{figures/prometheus_demo.pdf}
% \vspace{-4pt}
% \caption{Qualitative example from \emph{Prometheus}. The highest-ranked video
% frame gives the film's central threat greater immediacy than the IMDb still,
% which remains a strong second-ranked reference.}
% \label{fig:example-prometheus}
% \end{figure}

% \begin{figure}[!htbp]
% \centering
% \includegraphics[width=\linewidth]{figures/district_9_demo.pdf}
% \vspace{-4pt}
% \caption{Qualitative example from \emph{District 9}. The top video frame
% combines a human focal subject with the film's distinctive conflict and story
% world, while the IMDb still remains a strong second-ranked reference.}
% \label{fig:example-district9}
% \end{figure}

% \FloatBarrier

\section{Human Annotation Protocol}
\label{app:human-study}

The study set contains three films from each of the ten most frequent TMDB
genres, with no repeated film and at most two films from the same franchise.
For each film, we freeze one six-image study group containing one IMDb still
and five sampled frames. All six images are center-cropped and re-encoded as
unlabeled $1280\times720$ JPEGs with metadata removed, preventing image format
from revealing the IMDb still. Each group contributes all 15 unordered pairs,
yielding 30 groups and 450 human annotations. Pair order and left--right image
placement are randomized, and each pair receives one annotation.

Annotators see the title, year, synopsis, character names and portraits, and
key events while comparing each pair. They are asked which image would work
better as the movie's cover, thumbnail, or promotional still. Annotators select
with the arrow keys or by clicking an image, after which the interface saves
the choice and loads the next pair. Rubrics, benchmark rankings, image source,
and source identifiers are never shown. The interface supports English,
Chinese, Hindi, Turkish, and Persian. For agreement measurement, the VLM judge
evaluates each human-labeled pair in both image orders. A pair agrees only when
both VLM selections match the human preference.

\section{Additional Benchmark Statistics}
\label{app:statistics}

The 204-film corpus contains 128,085 LSMDC clips totaling 140.8 hours, with a
median source duration of 39.5 minutes per film and a median of 180 available
IMDb stills per film. The films span 1943--2014. The final release contains
2,031 reference chains and 12,186 chain images, comprising 2,031 IMDb stills
and 10,155 sampled frames. The 1,970 learned rubrics comprise 1,657 from round
one and 313 from round two. Another 70 rubrics are pruned with deleted anchors.
Each movie contains a mean of 9.66 rubrics (minimum 2, median 10, maximum 18).

The mean decisive rate is 0.953, and 52\% of chains are decisive on all 15
pairs. Anchors occupy ranks 1, 2, and 3 in 52.0\%, 25.8\%, and 22.2\% of
reference chains, respectively. Thus, construction favors the human-selected
IMDb still without forcing it to rank first.

\section{Construction Cost}
\label{app:cost}

Benchmark construction requires a median of 6,778 judge calls per film.
Constructing the 174 films not already covered by the human-study set requires
approximately 1.29 million calls. Full evaluation of the reported methods uses
221,370 local Qwen3.6-27B calls and incurs no API cost.